\documentclass[times, 10pt,twocolumn]{article} 
\usepackage{main}
\usepackage{times}
\usepackage{amsmath,graphicx}
\usepackage{epsfig,color}
\usepackage{url}
\usepackage{bigstrut,multirow,rotating}
\usepackage{textcomp,booktabs}
\usepackage{tabularx}
\usepackage{floatrow}
\newfloatcommand{capbtabbox}{table}[][\FBwidth]
\usepackage{caption}
\usepackage{colortbl}
\usepackage{xcolor}
\usepackage{graphicx}
\usepackage{multirow}
\usepackage{booktabs}

\usepackage{amsthm,amsmath,amssymb}
\usepackage{mathrsfs}
\begin{document}

\title{Enhancing General Face Forgery Detection via Vision Transformer with Low-Rank Adaptation}

\author{Chenqi Kong, Haoliang Li, and Shiqi Wang}

\maketitle
\thispagestyle{empty}

\begin{abstract}
Nowadays, forgery faces pose pressing security concerns over fake news, fraud, impersonation, etc. Despite the demonstrated success in intra-domain face forgery detection, existing detection methods lack generalization capability and tend to suffer from dramatic performance drops when deployed to unforeseen domains. To mitigate this issue, this paper designs a more general fake face detection model based on the vision transformer(ViT) architecture. In the training phase, the pretrained ViT weights are freezed, and only the Low-Rank Adaptation(LoRA) modules are updated. Additionally, the Single Center Loss(SCL) is applied to supervise the training process, further improving the generalization capability of the model. The proposed method achieves state-of-the-arts detection performances in both cross-manipulation and cross-dataset evaluations. 
\end{abstract}
\vspace{-0.6cm}
\Section{Introduction}
\vspace{-0.2cm}
With the rapid proliferation of digital face medias circulating on social media, non-expert attackers  can easily create fake face content due to unrestricted access to face media and the ease of implementing face manipulation techniques ($e.g.,$ Deepfakes \cite{hm16_20}, Face2Face \cite{dfcode}, FaceSwap \cite{thies2016face2face}, NeuralTextures \cite{thies2019deferred}, and other attacks \cite{yu2022towards, yu2023ba}.) \cite{kong2022digital}. Even worse, the availability of commercial tools and products (e.g., FakeApp \cite{fakeapp}) makes generating forgery faces much easier. The abuse of face manipulation have posed grand security concerns to the public at large, including fake news, financial fraud, identity theft, etc \cite{kong2021appearance}. Thus, it is of utmost importance to propose detection methods to counter the malicious attacks\cite{cai2020drl, cai2022learning} and build trust of digital facial medias. 

The past decades have witnessed significant progress in face forgery detection methodologies. Early works mainly focus on extracting handcrafted features such as lack of eye blinking~\cite{li2018ictu}, head pose inconsistency~\cite{yang2019exposing}, and face warping artifacts~\cite{li2018exposing} from the inputs. However, these methods suffer from limited accuracy and low generalization capability. Thanks to the advent of artificial intelligence and deep learning, many learning-based detection methods \cite{qian2020thinking, masi2020two, dang2020detection, li2020face, kong2022detect, li2022one} have been proposed and achieved outstanding detection performance under intra-domain settings. Nonetheless, learning-based methods are prone to overfitting to the training data, resulting in dramatic performance drops when deployed to unforeseen domains. In this vein, follow-up works such as \cite{li2021frequency, sun2021domain, shiohara2022detecting, luo2021generalizing} aim to mine more inherent and general artifacts from different manipulation techniques and datasets. Some multimodal-based models seek to use auxiliary modalities ($e.g.,$ audio modality) for more robust defense \cite{zhou2021joint, kong2022beyond, haliassos2022leveraging, kong2023m3fas}.

Inspired by the recent success of vision transformer (ViT) \cite{dosovitskiy2020image}, we apply the powerful ViT as the backbone of our framework. To achieve more general face forgery detection performance, we propose to incorporate the Low-Rank Adaptation (LoRA) \cite{hu2021lora} in this method. LoRA is a parameter-efficient tuning method that has been demonstrated effective in various domain generalization and few-shot learning tasks. Moreover, we borrow the idea of Single-Center Loss(SCL) \cite{li2021frequency} to make the features of real faces more compact and push the fake features away from the center of real features, thereby achieving more general face forgery detection. Compared with the ViT baseline, the designed model achieves a 6.6\% and 11.19\% AUC score boosts in challenging low-quality cross-manipulation and cross-dataset evaluations, respectively. 


\vspace{-0.2cm}
\Section{Related Work}
\vspace{-0.3cm}
 In this section, we first provide a broad review of prior literature on face forgery detection. Then, we briefly analyze and discuss typical parameter-efficient ViT tuning methods. 
\subsection{Face forgery detection methods}
\vspace{-0.1cm}
Early works on forgery face detection focused on extracting handcrafted features from the input face images/videos. Li $et.al.$ \cite{li2018ictu} analyzed the eye-blinking frequency to identify input authentication, while follow-up works \cite{yang2019exposing, li2018exposing} detected the head-pose inconsistency and face warping artifacts to determine the input face videos as real $v.s.$ fake. With the advent of artificial intelligence and deep learning, numerous learning-based methods have been proposed and achieved promising detection accuracy. Qian $et.al.$ \cite{qian2020thinking} propose to mine forgery-related features in frequency domain and achieved promising classification accuracy for low-quality fake videos. Dang $et.al.$ \cite{dang2020detection} proposed using manipulation region maps and applying an attention mechanism to achieve more accurate performance. Kong $et.al.$ \cite{kong2022detect} exploited both manipulation region and noise map to supervise the model training and obtain outstanding detection performance. Despite their demonstrated success in intra-domain evaluations, most existing face forgery detection methods lack generalization capability and cannot adapt well when deployed in unseen environments. To mitigate this issue, Face X-ray \cite{li2020face} proposed highlighting the boundary of manipulation regions in fake faces, thus achieving more general detection performance over different forgery techniques. SBI \cite{shiohara2022detecting} proposed a novel data augmentation method only using real face images to enforce the model to focus on inherent forgery artifacts rather than semantic contents. LTW \cite{sun2021domain} designed a more general model using meta learning and achieves outstanding cross-domain detection performance. This paper aims to achieve more general face forgery detection by taking advantage of LoRA modules and single-center loss to realize accurate and robust face manipulation detection.

\subsection{Parameter-efficient tuning for ViT}
In the past two years, vision transformers(ViT) have seen great success and have exploded into a plethora of vision applications, such as image classification, semantic segmentation, and object detection. Pretrained ViT models have also been widely used in downstream tasks and have achieved outstanding results through transfer learning. To improve the generalization capability of ViT and reduce computational cost, several parameter-efficient tuning methodologies have been proposed, such as Adapter \cite{houlsby2019parameter}, Low-Rank Adaptation (LoRA) \cite{hu2021lora}, and Visual Prompt Tuning (VPT) \cite{jia2022visual}. ViT Adapter is a neural network that includes a down-sample and up-sample layer, while LoRA optimizes the rank-decomposed changes of the two projection layers. VPT can be regarded as extra learnable input tokens in input space. Typically, these modules will be tuned and the ViT backbone parameters will be freezed with pretrained weights in the training phase. While the parameter-efficient tuning modules have been widely used in domain generalization and few-shot learning, how the tuning modules benefit the general forgery face detection has not been investigated yet. In this paper, we investigate the effectiveness of LoRA in forgery face detection and suggest incorporating Adapter and VPT in future works. 

\begin{figure}[h]
\centering
\includegraphics[scale=0.30]{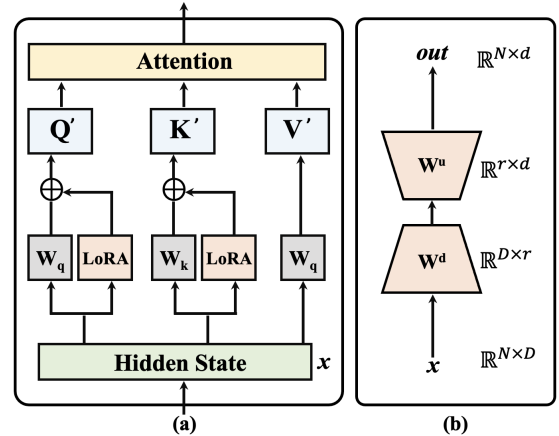}
\caption{(a). Illustration of the attention mechanism assembled with LoRA; (b). Details of LoRA.}
\label{LoRA}
\end{figure}

\Section{Proposed Method}
In this paper, we take the ViT \cite{dosovitskiy2020image} as our backbone. As shown in Fig.~\ref{LoRA}, we apply LoRA to the weights of query and key, in each attention layer. We fix the ViT with ImageNet weights and only update the LoRA parameters during the training process. The objective function of the model is the weighted summation of the cross-entropy loss $L_{ce}$ and the single-center loss $L_{scl}$: 
\begin{equation}
     L = L_{ce} + {\lambda} L_{scl},
\end{equation}
where $\lambda$ is the loss weight. The cross-entropy loss is detailed as:
\begin{equation}
     L_{ce}= -\frac{1}{N}\sum_{{i=1}}^{N}
     (c_{i}\log\hat{c}_{i}+(1-c_{i})\log(1-\hat{c}_{i})),
\end{equation}
where N is the number of input images. {$\hat{c}_{i}$} and {$c_{i}$} are the prediction result and ground-truth label. We dedicate the single-center loss $L_{scl}$ in Sec. 3.2.

\SubSection{Low-Rank Adaptation(LoRA)}
In typical attention mechanism, the query Q, key K, and value V can be obtained via Eqn.(3):
\begin{equation}
     Q=W_{q}x, K=W_{k}x, V=W_{v}x
\end{equation}
where $x \in \mathbb{R}^{N \times D}$,  $(W_{q}, W_{k}, W_{v}) \in \mathbb{R}^{D \times d}$. $W_{q}$, $W_{k}$, and $W_{v}$ are learnable weights.

\begin{figure}[h]
\centering
\includegraphics[scale=0.43]{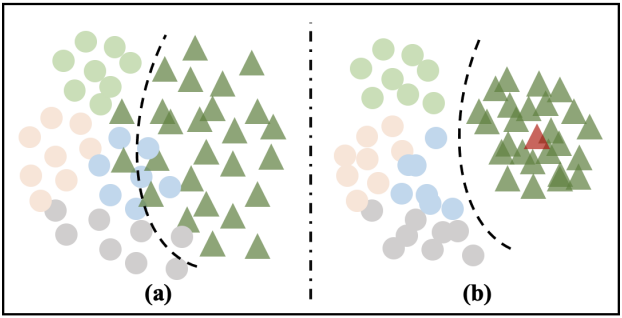}
\caption{Illustration of feature distribution (a). without single-center loss; (b). with single-center loss.}
\label{SCL}
\end{figure}

\begin{table*}
  \centering
  \scalebox{0.9}{\begin{tabular}{c|cc|cc|cc|cc|cc}
    \hline
    \textbf{Setting} & \multicolumn{2}{c|}{FF,FS,NT$\rightarrow$DF} & \multicolumn{2}{c|}{DF,FS,NT$\rightarrow$FF} & \multicolumn{2}{c|}{DF,FF,NT$\rightarrow$FS} & \multicolumn{2}{c|}{DF,FF,FS$\rightarrow$NT} & \multicolumn{2}{c}{\textbf{Average}}\\
    \hline	
    \textbf{Method} & AUC & ACC & AUC & ACC & AUC & ACC & AUC & ACC & AUC & ACC\\	
    \hline	 								
    ResNet18 \cite{he2016deep}  & 0.813 & 0.656 & 0.746 & 0.596 & 0.464 & 0.476 & 0.688 & 0.528 & 0.678 & 0.564 \\
    \hline	
    Xception \cite{chollet2017xception}  & 0.907 & 0.795 & 0.753 & 0.558 & 0.460 & 0.472 & 0.744 & 0.557 & 0.716 & 0.596 \\ 		
    \hline	 				
    EfficientNet \cite{tan2019efficientnet}  & 0.485 & 0.495 & 0.556 & 0.523 & 0.517 & 0.517 & 0.493 & 0.500 & 0.513 & 0.509 \\ 
    \hline	 				
    All-train EfficientNet \cite{tan2019efficientnet}  & 0.911 & 0.824 & 0.801 & 0.633 & 0.543 & 0.500 & 0.774 & 0.608  & 0.757 & 0.641 \\	 
    \hline	 				
    Focal-loss EfficientNet \cite{tan2019efficientnet}  & 0.903 & 0.813 & 0.798 & 0.608 & 0.503 & 0.484 & 0.759 &  0.604 & 0.741 & 0.627 \\ 
    \hline			
    Forensics Transfer \cite{cozzolino2018forensictransfer}  & N.A. & 0.720 & N.A. & 0.645 & N.A. & 0.460 & N.A. & 0.569 & N.A. & 0.599 \\ 	
    \hline					
    Multi-task \cite{nguyen2019multi}  & N.A. & 0.703 & N.A. & 0.587 & N.A. & 0.497 & N.A. & 0.603 & N.A. & 0.598 \\	
    \hline					
    MLDG \cite{li2018learning}  & 0.918 & 0.842 & 0.771 & 0.634 & 0.609 & 0.527 & \textbf{0.780} & 0.621 & 0.770 & 0.656 \\ 
    \hline	 								
    LTW \cite{sun2021domain}  & 0.927 & 0.856 & 0.802 & 0.656 & 0.640 & 0.549 & 0.773 & \textbf{0.653} & 0.786 & 0.679 \\ 
    \hline	
    \hline	
    ViT base \cite{dosovitskiy2020image}  & 0.771 & 0.701 & 0.656 & 0.582 & 0.510 & 0.498 & 0.554 & 0.517 & 0.623 &  0.575 \\		
    \hline					
    Ours & \textbf{0.935} & \textbf{0.862} & \textbf{0.875} & \textbf{0.753} & \textbf{0.651} & \textbf{0.554} & 0.707 & 0.626 & \textbf{0.792} & \textbf{0.699} \\		
    \hline				
  \end{tabular}}
  \caption{High quality (c23) cross-manipulation detection performance on unseen forgery methods.}
  \label{c23}
\end{table*}

In this model, we tune Low-Rank Adaptation(LoRA) modules, instead of $W_{q}$, $W_{k}$, and $W_{v}$, to obtain more general results. Fig.~\ref{LoRA} (a) illustrates the modified attention mechanism. $Q^{'}$, $K^{'}$, and $V^{'}$ can be obtained by Eqn.(4): 
\begin{equation}
     Q{'}=W_{q}x+sW_{q}^{d}W_{q}^{u}x; K{'}=W_{k}x+sW_{k}^{d}W_{k}^{u}x; V{'}=W_{v}x
\end{equation}
where $s$ is the fixed scale parameter. Fig.~\ref{LoRA} (b) shows the LoRA details. $W_{d} \in \mathbb{R}^{D \times r}$, $W_{u} \in \mathbb{R}^{r \times d}$, and $r$ is a hyper-parameter and generally much smaller than $d$ and $D$. 
By optimizing their rank-decomposed changes $W^{d}W^{u}$, we can benefit from the LoRA modules in following two aspects: (a). the proposed architecture is more computational-efficient since the number of the trainable parameters has been greatly reduced (going from $3Dd$ to $r(D+d)$); (b). the model retains abundant knowledge learned from ImageNet dataset and can be flexibly transferred to new tasks. 

\SubSection{Single-center loss(SCL)}
In this paper, we adopt the idea of single-center loss \cite{li2021frequency} to further improve the model's generalization capability. Fig.~\ref{SCL} illustrates the of feature distribution w/o and w/ SCL, where circles with different colors indicate different manipulation methods while triangles represent real samples. SCL is designed to make the feature distribution of real faces more compact and, at the same time, move fake features away from the center of real features (red triangle in Fig.~\ref{SCL} (b)). Eqn. (5) shows the single-center loss function:
\begin{equation} 
    L_{SCL}=d_{real}+max(d_{real}-d_{fake}+margin,0)
\end{equation}
where $d$ is the average distance between the real center and each feature, as shown in Eqn. (6):
\begin{equation} 
    d=\frac{1}{N}\sum_{{i=1}}^{N}||f_{i}-C||_{2}
\end{equation}
where $C$ represents the real center. We pick the features after the second last fully-connected layer to calculate $L_{SCL}$. By using such loss, the features of real and fake faces become more discriminative and separable, thus leading to a more general face forgery detection performance.
\Section{Experiments}
In this section, we conduct cross-domain experiments, including cross-manipulation and cross-dataset settings, to examine the robustness of the model. Then, we perform ablation studies to demonstrate the effectiveness of the LoRA module and the single-center loss.

\begin{table*}
  \centering
  \scalebox{0.9}{\begin{tabular}{c|cc|cc|cc|cc|cc}
    \hline
    \textbf{Setting} & \multicolumn{2}{c|}{FF,FS,NT$\rightarrow$DF} & \multicolumn{2}{c|}{DF,FS,NT$\rightarrow$FF} & \multicolumn{2}{c|}{DF,FF,NT$\rightarrow$FS} & \multicolumn{2}{c|}{DF,FF,FS$\rightarrow$NT} & \multicolumn{2}{c}{\textbf{Average}}\\
    \hline
    \textbf{Method} & AUC & ACC & AUC & ACC & AUC & ACC & AUC & ACC & AUC & ACC\\	
    \hline				
    ResNet18 \cite{he2016deep} & 0.741 & 0.673 & 0.648 & 0.600 & 0.634 & 0.594 & 0.598 & 0.567 & 0.655 & 0.609 \\
    \hline
    Xception \cite{chollet2017xception} & 0.766 & 0.694 & 0.696 & 0.643 & 0.626 & 0.593 & 0.597 & 0.552 & 0.671 & 0.621\\	
    \hline
    EfficientNet \cite{tan2019efficientnet} & 0.451 & 0.485 & 0.537  & 0.505& 0.512 & 0.503 & 0.499 & 0.497 & 0.500 & 0.498\\ 
    \hline
    All-train EfficientNet \cite{tan2019efficientnet} & 0.753 & 0.676 & 0.674 & 0.614 & 0.614 &  0.580 & 0.600 & 0.564 & 0.660 & 0.609 \\
    \hline
    Focal-loss EfficientNet \cite{tan2019efficientnet} & 0.749 & 0.674 & 0.672 & 0.610 & 0.596 & 0.575 & 0.605 & 0.566 & 0.656 & 0.606 \\ 				
    \hline
    Forensics Transfer \cite{cozzolino2018forensictransfer} & N.A. & 0.682 & N.A. & 0.550 & N.A. & 0.530 & N.A. & 0.550 & N.A. & 0.578 \\ 		
    \hline
    Multi-task \cite{nguyen2019multi} & N.A. & 0.667 & N.A. & 0.565 & N.A. & 0.517 & N.A. & 0.560 & N.A. & 0.577 \\	
    \hline
    MLDG \cite{li2018learning} & 0.730 & 0.671 & 0.617 & 0.581 & 0.617 &  0.581 & 0.607 & 0.569 & 0.643 & 0.601 \\
    \hline 				
    LTW \cite{sun2021domain} & 0.756 & 0.691 & \textbf{0.724} & \textbf{0.657} & 0.681 & 0.625 & \textbf{0.608} & \textbf{0.585} & 0.692 & 0.640\\
    \hline
    \hline
    ViT base \cite{dosovitskiy2020image} & 0.739 & 0.643 & 0.650 & 0.595 & 0.592 & 0.560 & 0.552 & 0.538 & 0.633 & 0.584 \\		
    \hline
    Ours & \textbf{0.818} & \textbf{0.735} & 0.686 & 0.638 & \textbf{0.710} &  \textbf{0.653} & 0.582 & 0.542 & \textbf{0.699} & \textbf{0.642} \\ 	
    \hline
  \end{tabular}}
  \caption{Low quality (c40) cross-manipulation detection performance on unseen forgery methods.}
  \label{c40}
\end{table*}
\subsection{Implementation details} 
The proposed framework is implemented by Pytorch. The model is trained using Adam optimizer with {$\beta_{1}$}=0.9 and {$\beta_{2}$}=0.999. We set the learning rate and weight decay as 1e-4 and 1e-5, respectively. The model is trained on 1 RTX 2080Ti GPUs with batch size 36. The FaceForensics++\cite{rossler2019faceforensics++} dataset is used as our training set. We follow the data split strategy in LTW \cite{sun2021domain} for fair comparison. 

\subsection{Evaluation on cross-manipulation detection}
FaceForensics++\cite{rossler2019faceforensics++} dataset provides 4000 fake videos generated by four manipulation techniques: Deepfake(DF), Face2Face(FF), FaceSwap(FS), and NeuralTextures(NT). Each video has three compression levels with different QPs: raw(QP=0), high quality(HQ QP=23), and low quality(LQ QP=40). To examine the generalization capability of the designed model on unseen manipulation techniques and accommodate real-world application scenarios, we conduct cross-manipulation evaluations on both HQ and LQ data, introduced next.

\noindent\textbf{Detection results on HQ data.} We apply leave-one-out cross-validation and average the results of four trials. Following prior arts, we report AUC and ACC scores in Table \ref{c23}. Compared with the baseline method (ViT base), the proposed method demonstrated a significant improvement over the baseline method (ViT base), with the average AUC score increasing from 0.623 to 0.792. This improvement can be attributed to the effectiveness of LoRA and SCL. 
On the other hand, our method is superior to the SOTA method LTW\cite{sun2021domain} in terms of the average detection performance.

\noindent\textbf{Detection results on LQ data.} Detecting low-quality manipulated faces is more challenging because severe compression can erase abundant forgery cues. Table \ref{c40} presents the detection results on the low-quality data. Compared to the ViT baseline, the average AUC and ACC scores of the proposed method get significant improvements: 6.6\% and 5.8\%, respectively. Additionally, our model achieves the best average detection performance, demonstrating its outstanding robustness under such a challenging setting.

\subsection{Evaluation on cross-dataset detection}
Evaluating the model on an unseen dataset is another practical scenario where the detection performance of most methods tends to degrade dramatically due to domain shift. In this paper, we train our model on FF++ Deepfake(both c23 and c40) subset and test it on unseen Deepfake datasets, including CelebDF \cite{li2020celeb}, DFD \cite{dfd}, DFDC \cite{dolhansky2020deepfake}, and Deepfake-TIMIT \cite{korshunov2018deepfakes}. The AUC detection scores are reported in Table~\ref{eval_crossdataset}, we can readily observe that the proposed method obtains 11.19\% AUC boost compared to the ViT baseline, demonstrating our model's generalization capability from another point of view.

\subsection{Ablation study}
To validate the effectiveness of LoRA and SCL in the task of general face forgery detection, we conduct an ablation evaluation under the challenging LQ cross-manipulation setting. As shown in Table \ref{Ablation}, the usage of LoRA modules significantly improves the AUC and ACC scores, and the SCL further boosts the low-quality cross-manipulation detection performance.

\begin{table*}
  \caption{Cross-dataset evaluation results.}
  \label{eval_crossdataset}
  \centering
  \renewcommand\arraystretch{1.15}
  \scalebox{0.8}{\begin{tabular}{c|c|c|c|c|c|c|c}
    \hline
     \textbf{Dataset} & CelebDF  & DFD (HQ) & DFD (LQ) & DFDC & DFMIT (HQ) & DFMIT (LQ) & \textbf{Average}\\
    \hline
    MesoNet \cite{afchar2018mesonet} & 58.85& 62.07 & 52.25 & 54.60 & 33.61 & 45.08 & 51.08 \\
    \hline
    MesoIncep4 \cite{afchar2018mesonet} & 68.26 & 79.18 & 63.27 & 61.92 & 16.12 & 27.47 & 52.70\\
    \hline
    ResNet50 \cite{he2016deep} & 67.09 & 69.60 & 60.61 & 61.97 & 41.95 & 47.27 & 58.08 \\
    \hline
    Face X-ray \cite{li2020face} & 71.89 & 69.61 & 62.89 & 58.97 & 42.52 & 50.05 & 59.32\\
    \hline
    DFFD \cite{dang2020detection} & 69.55 & 71.69 & 60.60  & 59.72 & 32.91 & 39.32 & 55.63 \\
    \hline
    Multi-task \cite{nguyen2019multi} & 65.18 & 70.75 & 58.61 & 57.38 & 16.53 & 15.59 & 47.34 \\
    \hline
    EfficientNet \cite{tan2019efficientnet} & 75.90 & 80.63  & 64.19 & 66.39 & 29.12 & 28.34 & 57.43 \\
    \hline
    F$^3$Net \cite{qian2020thinking} & 72.28 & 72.92 & 58.89 & 63.33 & 38.55 & 45.67 & 58.61 \\
    \hline
    Xception \cite{chollet2017xception} & 67.75 & 72.45 & 59.73 & 63.12 & 33.82 & 40.79 & 56.28 \\
    \hline
    D$\&$L (Effi.) \cite{kong2022detect} & 67.15 & 73.52 & 67.21 & 60.32 & 49.90 & 51.01 & 61.52\\
    \hline
    D$\&$L (Xcep.) \cite{kong2022detect} & 70.65 & 76.23 & 64.53 & 63.31 & 47.20 & \textbf{56.08} & 63.00 \\
    \hline
    ViT-base \cite{dosovitskiy2020image} & 71.23 & 61.32 & 59.51 & 66.84 & 56.69 & 47.41 & 60.50 \\
    \hline
    Ours & \textbf{83.76} & \textbf{83.42} & \textbf{68.31} & \textbf{71.74} & \textbf{70.36} & 52.52 & \textbf{71.69} \\ 
    \hline 						
\end{tabular}}
\end{table*}

\begin{table}
  \caption{Ablation study (c40 cross-manipulation).}
  \label{Ablation}
  \centering
  \renewcommand\arraystretch{1.15}
  \scalebox{1.0}{\begin{tabular}{c|c|c|c|c}
    \hline
    ViT & LoRA & SCL & AUC & ACC \\
    \hline
    $\surd$ & - & -  & 0.633 & 0.584\\
    \hline
    $\surd$ & $\surd$ & -  & 0.684 & 0.630\\
    \hline
    $\surd$ & $\surd$ & $\surd$  & \textbf{0.699} & \textbf{0.642}\\
    \hline
\end{tabular}}
\end{table}
\Section{Conclusions}
In this paper, we presented a general and robust face forgery detection method based on ViT backbone. Firstly, the backbone is initialized with ImageNet weights, and the loaded parameters are frozen during the training process. Then, we tune the LoRA modules under the joint supervision of cross-entropy and single center losses. By doing this, the number of trainable parameters can be greatly reduced and much computational resource can be saved. Extensive experiments demonstrate that the use of LoRA and SCL can improve the generalization capability of the forgery detection model. The proposed method can serve as a basis for developing ViT-based face forgery detection models.

\section{Acknowledgement}
This work is supported in part by Shenzhen Virtural University Park, The Science Technology and Innovation Committee of Shenzhen Municipality (Project No: 2021Szvup128).

This work is also supported by the Research Grant Council (RGC) of Hong Kong through Early Career Scheme (ECS) under the Grant 21200522 and Sichuan Science and Technology Program 2022NSFSC0551.

\bibliographystyle{main}
\bibliography{main}

\end{document}